\documentclass[10pt]{article} 
\usepackage[preprint]{rlc}

\usepackage{amssymb}            
\usepackage{mathtools}          
\usepackage{mathrsfs}           
\usepackage{graphicx}           
\usepackage{subcaption}         
\usepackage[space]{grffile}     
\usepackage{url}                

\usepackage{algorithm}
\usepackage{algpseudocode}

\definecolor{enccolor}{RGB}{83, 151, 139}
\definecolor{zcolor}{RGB}{235, 197, 82}
\definecolor{scolor}{RGB}{235, 159, 82}
\definecolor{hcolor}{RGB}{235, 133, 133}
\definecolor{deccolor}{RGB}{108, 108, 108}
\definecolor{acolor}{RGB}{133, 184, 235}
\definecolor{rcolor}{RGB}{158, 82, 235}
\definecolor{ccolor}{RGB}{82, 82, 235}
\definecolor{vcolor}{RGB}{31, 133, 235}
\definecolor{g1}{gray}{0.3}
\definecolor{g2}{gray}{0.5}
\definecolor{g3}{gray}{0.7}
\definecolor{bluechart}{rgb}{0.00392156862745098, 0.45098039215686275, 0.6980392156862745}
\definecolor{lightbluechart}{rgb}{0.33725490196078434, 0.7058823529411765, 0.9137254901960784}
\definecolor{greenchart}{rgb}{0.00784313725490196, 0.6196078431372549, 0.45098039215686275}

\title{Do Agents Dream of Electric Sheep?: Improving Generalization in Reinforcement Learning through Generative Learning}

\author{Giorgio Franceschelli  \\
    giorgio.franceschelli@unibo.it \\
    Department of Computer Science and Engineering\\
    University of Bologna, Italy
    \And
    Mirco Musolesi \\
    m.musolesi@ucl.ac.uk\\
    Department of Computer Science \\
    University College London, United Kingdom\\
    Department of Computer Science and Engineering\\
    University of Bologna, Italy}

\begin{document}

\maketitle

\begin{abstract}
The \textit{Overfitted Brain} hypothesis \citep{hoel21} suggests dreams happen to allow generalization in the human brain. Here, we ask if the same is true for reinforcement learning agents as well. Given limited experience in a real environment, we use imagination-based reinforcement learning to train a policy on \textit{dream}-like episodes, where non-imaginative, predicted trajectories are modified through generative augmentations. Experiments on four ProcGen environments show that, compared to classic imagination and offline training on collected experience, our method can reach a higher level of generalization when dealing with sparsely rewarded environments.
\end{abstract}

\section{Introduction}

Deep Reinforcement Learning (RL) has emerged as a very effective mechanism for dealing with complex and intractable AI tasks of different nature. Model-free methods that essentially learn by trial and error have solved challenging games \citep{mnih15}, performed simulated physics tasks \citep{lillicrap16}, and aligned large language models with human values \citep{ouyang22}. However, RL commonly requires an incredible amount of collected experience, especially compared to the one required by humans \citep{tsividis17}, limiting its applications to real-world tasks.
 
Model-based RL \citep{sutton17} constitutes a promising direction towards sample efficiency. Learning a world model capable of predicting the next states and rewards conditioned on actions allows the agent to plan \citep{schrittwieser20} or build additional training trajectories \citep{ha18}. In particular, recent imagination-based methods \citep{hafner20a, hafner21, hafner23, micheli23} have shown remarkable performance simply by learning from imagined episodes within a learned latent space. Such imagined trajectories are commonly mentioned as \textit{dreams}. However, these dreams are nothing like \textit{human} dreams, as they essentially try to mimic reality as best as possible.
According to the \textit{Overfitted Brain} hypothesis \citep{hoel21}, dreams happen to allow generalization in the human brain. In particular, it is by providing hallucinatory and corrupted content \citep{hoel19} that are far from the limited daily experiences (i.e., the training set) that dreaming helps prevent overfitting. We build on this intuition and ask: \textit{can human-like ``dreams'' help RL agents generalize better when dealing with limited experience?}

In this paper, we explore whether this type of experience augmentation based on dream-like generated trajectories helps generalization and, consequently, improves learning. In particular, we consider the situation in which only a limited amount of real experience (analogously to ``daylight activities'' for humans) is available, and we question whether building a world model upon it and leveraging it to generate dream-like experiences improves the agent's generalization capabilities. To simulate the hallucinatory and corrupted nature of dreams, we propose to transform the classic imagined trajectories with \textit{generative augmentations}, i.e., through interpolation with random noise \citep{wang20}, DeepDream \citep{mordvintsev15}, or critic's return optimization (similar to class visualization; \cite{simonyan14}). We evaluate them on four ProcGen environments \citep{cobbe20}, a standard suite for generalization in RL \citep{kirk23}. Our experiments\footnote{We plan to release the code soon.} show that for sparsely rewarded environments our method can reach higher levels of generalization compared with \textit{classic} imagination and offline training.

The main contributions of this paper can be summarized as follows:

\begin{itemize}
    \item We leverage existing world models (Section \ref{preliminaries}) learned from limited data to construct imagined trajectories from randomly generated states.
    \item We define three novel types of experience augmentation based on hallucination and corruption of the trajectories to improve generalization, making them closer to human-like dreams in a sense (Section \ref{method}).
    \item We evaluate the generalization capabilities of our methods against standard imagination and offline training over collected experience using ProcGen, showing how dream-like trajectories can help generalize better (Section \ref{experiments}). 
\end{itemize}

\section{Related Work} \label{relatedwork}

\subsection{Imagination-Based RL} \label{imaginationbased}

World models were first introduced by Dyna \citep{sutton91} and then extensively studied in model-based RL for planning \citep{chua18,gal16,hafner19,henaff17,schrittwieser20}. It has been shown that they can also help model-free agents by reducing their state-space dimensionality \citep{banijamali18,watter15}, by guiding their decisions through the provision of additional information \citep{buesing18,racaniere17}, and by constructing imagined trajectories to be used in place of real (expensive) experience. While it is possible to directly work on highly-dimensional observations \citep{kaiser20}, the main line of research consists of learning a compact latent representation of the environment with a posterior model to encode current observation and a prior model to predict the encoded state. The prior model is then used to construct imagined trajectories and can be implemented with \citep{ha18,hafner19,hafner20a} or without \citep{lee20a,zhang19} a recurrent layer that keeps track of the episode history. \cite{igl18} also train the world model on the agent objective. Instead of working on states, \cite{nagabandi18} model the environment's dynamics through the difference between consecutive states. \cite{hafner21,hafner23} replace the classic continuous latent space with discrete representations. \citet{sekar20} use an intrinsic reward based on ensemble disagreement to guide imagined exploration. \cite{zhu20} employ latent overshooting to train the dynamics-agent pair together. \cite{mu21} construct imagined trajectories not only from real states but also from derived states whose features are randomly modified. Transformers \citep{vaswani17} can be used to represent and learn the world model in place of the recurrent layer \citep{chen22} or all predictive components \citep{micheli23}. In general, all these methods only produce trajectories that adhere as close as possible to the real ones, lacking the \textit{divergent} aspect of dreaming that helps humans avoid overfitting.

\subsection{Generalization in RL} \label{generalization}

Different techniques have been proposed to approach generalization in RL. A first strategy is to learn an environment representation decoupled from the specific policy optimization \citep{jaderberg17,stooke21}. Another is to adopt techniques used in supervised learning to avoid overfitting, e.g., dropout, batch normalization, and specific convolutional architectures \citep{cobbe19,farebrother18,igl19}. An alternative is to improve the agent's architecture, the training process, or the experience replay sampling technique \citep{jiang21}. For example, \cite{raileanu21b} propose to train the value function with an auxiliary loss that encourages the model to be invariant to task-irrelevant properties. Also, post-training distillation may help improve generalization to new data \citep{lyle22}, as well as learning an embedding in which states are close when their optimal policies are similar \citep{agarwal21} or interpolating between collected observations \citep{wang20}. Another strategy is to use data augmentation to increase the quantity and variability of training data \citep{cobbe19,laskin20,lee20b,tobin17,yarats21,ye20}. The most appropriate augmentation technique can even be learned and not selected a priori \citep{raileanu21a}. Finally, \cite{ghosh21} propose to deal with the epistemic uncertainty introduced by generalization through an ensemble-based technique, with multiple policies trained on different subsets of the distribution and then combined.
Our approach is conceptually close to data augmentation, but with augmentations based on the learned world model itself, thus providing semantically richer transformations.

\section{Preliminaries} \label{preliminaries}

\subsection{Modeling Latent Dynamics} \label{world}

World models represent a compact and learned version of the environment capable of predicting imagined future trajectories \citep{sutton91}. When the inputs are high-dimensional observations $o_t$ (i.e., images), Dreamer \citep{hafner20a, hafner21, hafner23} represents the current state of the art due to its ability to learn compact latent states $z_t$. In general, Dreamer world model consists of the following components:
\begin{align*}
        &\text{Recurrent model:} & &h_t = f_{\theta}\!\left(h_{t-1}, z_{t-1}, a_{t-1}\right) \\
        &\text{Encoder model:} & &z_t \sim q_{\theta}\!\left(z_t | h_t, o_t\right) \\
        &\text{Transition predictor:} & &\hat{z}_t \sim p_{\theta}\!\left(\hat{z}_t | h_t\right) \\
        &\text{Reward predictor:} & &\hat{r}_t \sim p_{\theta}\!\left(\hat{r}_t | h_t, z_t\right) \\
        &\text{Continue predictor:} & &\hat{c}_t \sim p_{\theta}\!\left(\hat{c}_t | h_t, z_t\right) \\
        &\text{Decoder model:} & &\hat{o}_t \sim p_{\theta}\!\left(\hat{o}_t | h_t, z_t\right) \\
\end{align*}
The deterministic recurrent state $h_t$ is predicted by a Gated Recurrent Unit (GRU) \citep{cho14}, while the encoder and decoder models use convolutional neural networks for visual observations. Overall, the Recurrent State-Space Model \citep{hafner19}, an architecture that contains recurrent, encoder, and transition components, learns to predict the next state only from the current one and the action, while also allowing for correct reward, continuation bit, and image reconstructions.

While $z_t$ was originally parameterized through a multivariate normal distribution, more recent works \citep{hafner21,hafner23} consider a discrete latent state. In particular, they use a vector of $C$ one-hot encoded categorical variables (i.e., a very sparse binary vector). \cite{hafner23} parameterize this categorical distribution as a mixture of 1\% uniform and 99\% neural network output. Moreover, instead of regressing the rewards via squared error, they propose a learning scheme based on two transformations: first, the rewards are symlog-transformed \citep{webber13}; then, they are two-hot encoded, i.e., converted into a vector of $K$ values where $K-2$ are 0, and the remaining, consecutive two are positive weights whose sum is 1. The $K$ values correspond to equally spaced buckets, so that by multiplying the vector with the bucket values we reconstruct the original reward. We adopt this solution: this helps learning, especially in environments with very sparse rewards.

Overall, given a sequence of inputs $\{o_{0:T-1}, a_{0:T-1}, r_{1:T}, c_{1:T}\}$, the world model is trained to minimize the following loss:

\begin{equation} \label{eq:worldloss}
    \mathcal{L}\!\left(\theta\right) = \mathbb{E}_{q_{\theta}}\!\left[\sum_{t=1}^T\!\left(\mathcal{L}_{pred}\!\left(\theta\right) + \beta_1 \mathcal{L}_{dyn}\!\left(\theta\right) + \beta_2 \mathcal{L}_{rep}\!\left(\theta\right)\right)\right]
\end{equation}

where $\mathcal{L}_{pred}$ trains the decoder model via mean squared error loss, the reward predictor via categorical cross-entropy loss, and the continue predictor via binary cross-entropy loss; while $\mathcal{L}_{dyn}$ and $\mathcal{L}_{rep}$ consider the same Kullback-Leibler (KL) divergence between $q_{\theta}\!\left(z_t | h_t, o_t\right)$ and $p_{\theta}\!\left(\hat{z}_t | h_t\right)$, but using the stop-gradient operator on the former for the first loss and on the latter for the second loss. Moreover, free bits \citep{kingma16} are employed to clip the KL divergence below the value of 1 nat. Finally, $\beta_1$ and $\beta_2$ are scaling factors necessary to encourage learning an accurate prior over increasing posterior entropy \citep{hafner21}.

\subsection{Learning through Imagination} \label{imagination}

In general, leveraging the world model detailed in Section \ref{world}, a policy $\pi_{\phi}\!\left(a_t | s_t\right)$ can be learned by acting only in the latent space of imagination: given a compact latent state $\hat{s}_t^{im} = \left(h_t^{im}, \hat{z}^{im}_t\right)$, the agent selects action $a_t$, returns it to the world model, and receives $\hat{r}_{t+1}$, $\hat{c}_{t+1}$, and $\hat{s}^{im}_{t+1}$. Furthermore, a critic $v_{\phi}\!\left(v_t | s_t\right)$ is simultaneously learned to predict the state-value function $v_t$. This process is repeated until a fixed imagination horizon is reached and the policy can be learned from the imagined experience as it would have done by acting in the real environment. The agent can be trained on the collected trajectories either by direct reward optimization (leveraging the differentiability of the trajectory construction and back-propagating through the reward model; \cite{hafner20a}) or by using a model-free policy gradient method, e.g., REINFORCE \citep{williams92}.

\section{Dream to Generalize} \label{method}

By starting from the models presented in Section \ref{preliminaries}, our method proposes first to learn a latent world model from real experience; to augment the imagined trajectories to resemble human dreams (Section \ref{generatingdreams}); and then to exploit such new trajectories to learn policies that are more keen to generalize (Section \ref{policylearning}).

\subsection{Generating Human-Like Dreams} \label{generatingdreams}

\begin{figure}[ht]
  \centering
  \includegraphics[width=.8\linewidth]{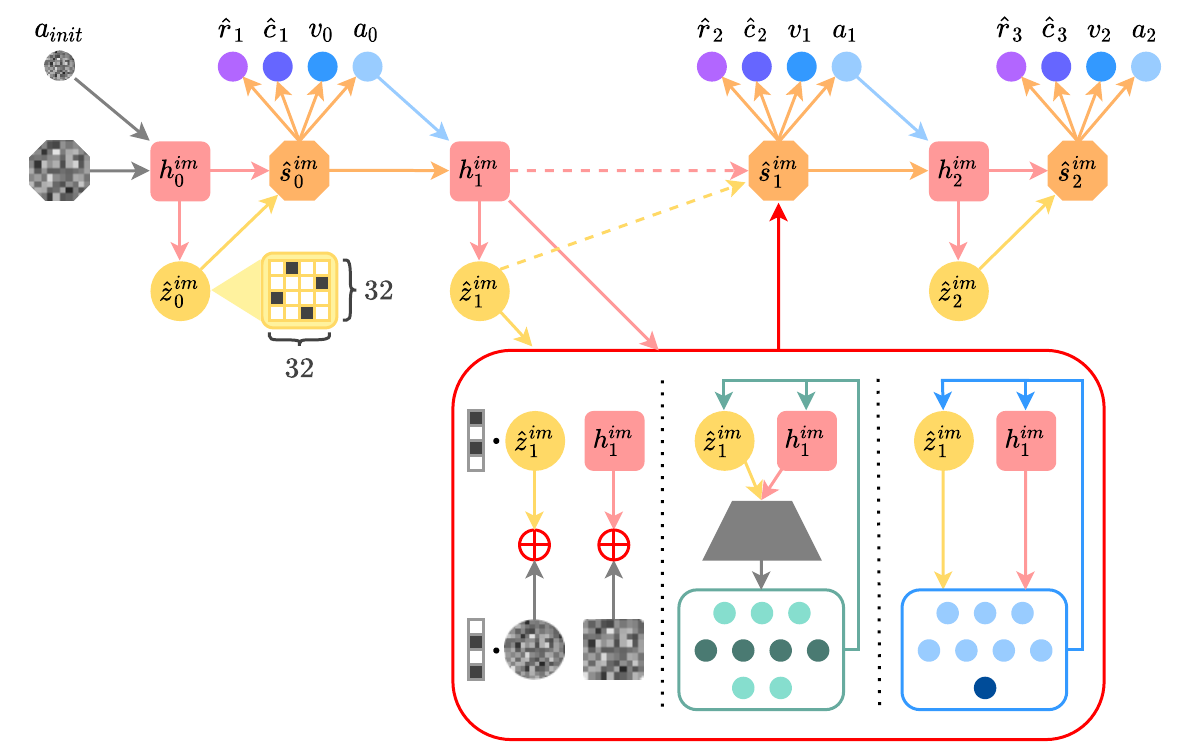}
  \caption[Dreaming process]{At imagination time, we start from a \textcolor{g3}{\textbf{ran}}\textcolor{g1}{\textbf{dom}} \textcolor{g2}{\textbf{lat}}\textcolor{g1}{\textbf{ent}} \textcolor{g3}{\textbf{sta}}\textcolor{g2}{\textbf{te}} and then we only leverage the predicting capabilities of our world model to obtain future \textcolor{scolor}{\textbf{latent states}} (the concatenation of a \textcolor{zcolor}{\textbf{discrete latent vector}} and a \textcolor{hcolor}{\textbf{recurrent hidden state}}), \textcolor{rcolor}{\textbf{rewards}} and \textcolor{ccolor}{\textbf{termination bits}} given the \textcolor{acolor}{\textbf{actions}} from the agent. To introduce a \textcolor{red}{\textbf{dream-like transformation}}, we modify the current \textcolor{scolor}{\textbf{latent state}} with a small probability by doing one of three operations: \textit{interpolate} it with \textcolor{g2}{\textbf{ran}}\textcolor{g3}{\textbf{dom}} \textcolor{g1}{\textbf{noi}}\textcolor{g2}{\textbf{se}}; \textit{DeepDream} its corresponding observation from the \textcolor{deccolor}{\textbf{decoder}} by maximizing the activation of the \textcolor{enccolor}{\textbf{encoder}} last convolution layer; \textit{optimize} it to maximize the absolute value of \textcolor{vcolor}{\textbf{critic output}}.}
  \label{fig:dreaming}
\end{figure}

Given a trained world model, we can use it to construct imagined trajectories as detailed in Section \ref{imagination}. Crucially, instead of starting each trajectory from a real collected state (as is commonly done in the literature), we start from randomly generated states $\hat{s}_0^{im} = \left(h_0^{im}, \hat{z}_0^{im}\right)$ with

\begin{equation} \label{eq:initialstates}
\begin{split}
    & h_{init} \sim \mathcal{N}\!\left(0, I\right), \\
    & \hat{z}_{init} = \text{\textit{one-hot}}\!\left(u_{1:C}\right), u_c \sim \mathcal{U}\!\left(0, J-1\right) \text{ for } c = 1...C, \\
    & h_0^{im} = f_{\theta}\!\left(h_{init}, \hat{z}_{init}, a_{init}\right), \\
    & \hat{z}_0^{im} \sim p_{\theta}\!\left(h_0^{im}\right),
\end{split}
\end{equation}

\noindent where $J$ is the number of classes each of the $C$ categorical variables can assume, $\text{\textit{one-hot}}\!\left(\cdot\right)$ transforms a list of categorical variables into a vector of one-hot encoded vectors, and $a_{init}$ is a zero vector.

In addition, to obtain more human-like dreams, we leverage the world model to propose three perturbation strategies (see Figure \ref{fig:dreaming} for a summary of the process):

\begin{itemize}
    \item \textbf{Random swing}, i.e., interpolation between the current state $\hat{s}^{im}_t = \left(h^{im}_t, \hat{z}^{im}_t\right)$ and a random noise state (similar to \cite{wang20}). 
    In particular, we perturb the hidden state $h^{im}_t$ by adding a random vector $h_{rand} \sim \mathcal{N}\!\left(0, I\right)$. Instead, our transformation over the latent state $\hat{z}^{im}_t$ can be formalized as:

\begin{equation} \label{eq:randomjumps}
\begin{split}
    &\hat{z}^{im}_t = \text{\textit{one-hot}}\!\left(\lambda \cdot \text{\textit{reverse-one-hot}}\!\left(\hat{z}^{im}_t\right) + \left(1 - \lambda \right) \cdot u_{1:C}\right), \\
    &\lambda \sim \mathrm{Bin}\!\left(C, p_{swing}\right), \\
    &u_c \sim \mathcal{U}\!\left(0, J-1\right) \text{ for } c = 1...C
\end{split}
\end{equation}

    where $\text{\textit{reverse-one-hot}}\!\left(\cdot\right)$ inverts the one-hot encoding, i.e., recovers the list of categorical variables, and $p_{swing} = 0.5$ is the probability of making a \textit{swing}. In other words, each categorical variable is changed into a randomly sampled class with probability $p_{swing}$.
    This simulates the corruption of dream content and the sudden visual changes we commonly experience during REM sleep \citep{andrillon15}. 

    \item \textbf{DeepDream}, i.e., by iteratively adjusting the image reconstructed from the state to maximize the firing of a model layer \citep{mordvintsev15}. Specifically, we consider the last convolutional layer of the encoder, which should learn the building elements of real images. Given $q_{\theta}^{LC}\!\left(\cdot\right)$ as the activation of the last convolutional layer of dimension D, we transform the hidden state $h^{im}_t$ and the latent state $\hat{z}^{im}_t$ via gradient ascent over the following objective:

    \begin{equation}
        g_{dd} = \nabla_{h^{im}_t, \hat{z}^{im}_t} 
        \dfrac{\sum _{i=1}^D{q_{\theta}^{LC}}\!\left(p_{\theta}\!\left(h^{im}_t, \hat{z}^{im}_t\right)\right)_i}{D}.
    \end{equation}

    This simulates the hallucinatory nature of dreams.

    \item \textbf{Value diversification}, i.e., by iteratively adjusting the state $\hat{s}^{im}_t$ to maximize the squared difference between the value of the critic prediction at iteration $\tau$ and iteration $0$. We perform a gradient ascent over the following objective:

    \begin{equation}
        g_{vd} = \nabla_{h^{im}_t, \hat{z}^{im}_t} \left(v_{\phi}\!\left(h^{im}_t, \hat{z}^{im}_t\right) - v_{\phi}(h^{inp}_t, \hat{z}^{inp}_t)\right)^2,
    \end{equation}

    where $\hat{s}^{inp}_t = (h^{inp}_t, \hat{z}^{inp}_t)$ is the state before optimization. The squared difference is considered to optimize for both positive and negative changes in the critic's prediction. In addition, at each iteration, $\hat{z}^{im}_t$ is transformed to keep it as a vector of one-hot categorical variables.  
    The value diversification transformation suddenly introduces or removes goals or obstacles, simulating the narrative content and the fact that dreams commonly resemble daily aspects that are significant to us, especially threatening events \citep{revonsuo00}. In fact, simulating negative experiences might allow an agent to learn what to avoid in practice.
    
\end{itemize}

\begin{figure}[t]
     \centering
     \begin{subfigure}[b]{0.19\textwidth}
         \centering
             \includegraphics[width=\textwidth]{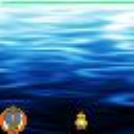}
         \caption{\small{Original.}}
         \label{fig:real}
     \end{subfigure}
     \hfill
     \begin{subfigure}[b]{0.19\textwidth}
         \centering
             \includegraphics[width=\textwidth]{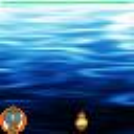}
         \caption{Reconstruction.}
         \label{fig:id}
     \end{subfigure}
     \hfill
     \begin{subfigure}[b]{0.19\textwidth}
         \centering
             \includegraphics[width=\textwidth]{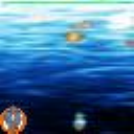}
         \caption{Random swing.}
         \label{fig:random_jump}
     \end{subfigure}
     \hfill
     \begin{subfigure}[b]{0.19\textwidth}
         \centering
             \includegraphics[width=\textwidth]{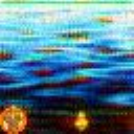}
         \caption{DeepDream.}
         \label{fig:deepdream}
     \end{subfigure}
     \hfill
     \begin{subfigure}[b]{0.19\textwidth}
         \centering
             \includegraphics[width=\textwidth]{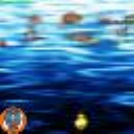}
         \caption{Value div.}
         \label{fig:reward_dream}
     \end{subfigure}
     \hfill
        \caption{An example of the three generative augmentations on a state from Plunder environment.}
        \label{fig:dreams}
\end{figure}

Figure \ref{fig:dreams} reports a visual example of the three transformations.
%
We alter each state $\hat{s}^{im}_t$ with a small probability $\epsilon_{dream} = \frac{1}{H}$ with $H$ imagination horizon. In this way, each trajectory includes, on average, one transformed state.

\subsection{Learning by Day and Night} \label{policylearning}

Our method can be divided into two stages. During the first, our agent plays a limited number of real episodes (the \textit{day} experience), which are used to train both the world model and the agent in an E2C-like setting \citep{watter15}, where the agent receives the encoding of the real observation by the world model as the state. We then leverage the world model to generate additional dreamed episodes (the \textit{night} experience), which are used to keep on training the agent. Algorithm \ref{alg:training} summarizes the entire learning process.

\begin{algorithm}[ht]
\caption{Learning to generalize by day and by night}\label{alg:training}
\begin{algorithmic}
\State \textbf{Require} $S$ number of seed episodes, $E_d$ day epochs, $E_n$ night epochs, $U_w$ update steps per day epoch, $U_a$ update steps per night epoch, $B_w$ world batch size, $B_a$ agent batch size, $L$ sequence length, $H$ imagination horizon, $T_d$ steps in environment per day epoch.
\State \textbf{Initialize} neural network parameters $\theta$ and $\phi$ randomly.
\State \textbf{Initialize} dataset $\mathcal{D}$ with $S$ random seed episodes.
\State $o_1 \leftarrow \mathtt{env.reset}\!\left(\right)$
\For{day epoch $e_d = 1...E_d$} \Comment Day experience
\For{update step $u = 1...U_w$}
\State Draw $B_w$ data sequences $\left\{\left(o_t,a_t,r_{t+1},c_{t+1}\right)\right\}_{t=k}^{k+L} \sim \mathcal{D}$.
\State Update $\theta$ through representation learning (Equation \ref{eq:worldloss}).
\EndFor
\For{day step $t = 1...T_d$}
\State Compute $s_t = \left(h_t,z_t\right)$, $h_t = f_{\theta}\!\left(h_{t-1}, z_{t-1}, a_{t-1}\right)$, $z_t \sim q_{\theta}\!\left(h_t, o_t\right)$.
\State Compute $a_t \sim \pi_{\phi}\!\left(a_t | s_t\right)$.
\State $o_{t+1}, r_{t+1}, c_{t+1} \leftarrow \mathtt{env.step}\!\left(a_t\right)$.
\EndFor
\State Update $\phi$ through PPO (Equation \ref{eq:ppo}) using collected experience.
\State Add collected experience to dataset $\mathcal{D} \leftarrow \mathcal{D} \cup \left\{\left(o_t, a_t, r_{t+1}, c_{t+1}\right)_{t=0}^{T_d-1}\right\}$.
\State Evaluate $\pi_{\phi}$ on $\mathtt{test\_env}$.
\EndFor
\For{night epoch $e_n = 1...E_n$} \Comment Night experience
\State Sample $B_a \cdot U_a$ random states $\hat{s}_0^{im}$ according to Equation \ref{eq:initialstates}.
\State Dream trajectories $\left\{\left(\hat{s}^{im}_{\tau}, a_{\tau}, \hat{r}_{\tau+1}, \hat{c}_{\tau+1}\right)\right\}_{\tau=0}^{H-1}$ from each $\hat{s}_0^{im}$.
\State Update $\phi$ through PPO (Equation \ref{eq:ppo}) using generated experience.
\State Evaluate $\pi_{\phi}$ on $\mathtt{test\_env}$.
\EndFor
\end{algorithmic}
\end{algorithm}

The latent world model is trained as detailed in Section \ref{world}. As far as the agent is concerned, following \cite{hafner23}, we adopt an actor-critic architecture that works on the latent state $s_t = \left(h_t, z_t\right)$. However, instead of using REINFORCE, we train it during both \textit{day} and \textit{night} stages through Proximal Policy Optimization (PPO; \cite{schulman17}), which we find helpful to obtain a more stable training. The overall loss is defined as follows:

\begin{equation} \label{eq:ppo}
    L_t\!\left(\phi\right) = \sum_{t=0}^{T-1} \mathbb{E}_{\pi_{\phi}, v_{\phi}, p_{\theta}} \!\left[ - L^{CLIP}_t\!\left(\phi\right) + c_v L^{VF}_t\!\left(\phi\right) - c_e \mathrm{H}\!\left[\pi_{\phi}\right]\!\left(s_t\right) \right],
\end{equation}

\noindent where

\begin{equation} \label{eq:clip}
    L^{CLIP}_t\!\left(\phi\right) = \hat{\mathop{\mathbb{E}}}_t \!\left[ \min\!\left(\frac{\pi_{\phi}\!\left(a_t | s_t\right)}{\pi_{\phi{old}}\!\left(a_t | s_t\right)} \hat{A}_t, \text{ clip}\!\left(\frac{\pi_{\phi}\!\left(a_t | s_t\right)}{\pi_{\phi{old}}\!\left(a_t | s_t\right)}, 1 - \epsilon, 1 + \epsilon\right) \hat{A}_t\right)  \right]
\end{equation}

\noindent is the clipped surrogate objective that modifies the policy in the right direction while preventing large changes, and $\mathrm{H}\!\left[\pi_{\phi}\right]$ is an entropy bonus scaled by $c_e$ coefficient. Instead, the value function is modeled as the reward (see Section \ref{world}): the critic network $v_{\phi}$ produces a softmax distribution across bins, where each one represents a partition of the potential value range.
Therefore, its loss function is defined as follows:

\begin{equation} \label{eq:value}
    L^{VF}_t\!\left(\phi\right) = - \text{\textit{sg}}\!\left(\text{\textit{two-hot}}\!\left(\text{\textit{symlog}}\!\left(V^{target}_t\right)\right)\right)^T \ln\!\left(v_{\phi}\!\left(\cdot | s_t\right)\right),
\end{equation}

where $\text{\textit{sg}}\!\left(\cdot\right)$ stops the gradient, and $\text{\textit{two-hot}}\!\left(\cdot\right)$ and $\text{\textit{symlog}}\!\left(\cdot\right)$ transform the target discounted return $V^{target}_t = \hat{A}_t + v_{\phi}\!\left(s_t\right)$ into its two-hot encoded, symlog-transformed version. Finally, the advantage is estimated following \cite{schulman15}:

\begin{equation} \label{eq:adv}
    \hat{A}_t = \sum_{i=0}^{T-t} \left(\gamma\lambda\right)^i \left( r_{t+i+1} + \gamma v_{\phi}\!\left(s_{t+i+1}\right) - v_{\phi}\!\left(s_{t+i}\right) \right),
\end{equation}

and normalized according to the scheme proposed in \cite{hafner23}, i.e., by dividing it by $\max\!\left(1, P\right)$ with $P$ as the exponentially decaying average of the range from their $5^{th}$ to their $95^{th}$ batch percentile. This improves exploration under sparse rewards. Finally, we also adopt two additional mechanisms to improve the learning process: we penalize non-successful completion of the tasks in sparsely rewarded environments, i.e., we associate a negative reward to a non-positive termination state; and we prioritize sampling of non-zero rewarded sequences during training, which helps learn a meaningful reward predictor.

\section{Experiments} \label{experiments}

In the following, we present experiments on the generalization capabilities of our proposed approach using ProcGen  \citep{cobbe20}, a simple yet rich set of environments for RL generalization evaluation.

\subsection{Setup}

ProcGen is a suite of 16 procedurally generated game-like environments. To benchmark the generalization capabilities of our approach, only a small subset of the distribution of levels ($N=200$) is used to train the agent and the full distribution to test it. Due to resource constraints, our experiments consider the ProcGen suite in easy mode; we limit the collected real experience to 1M steps, far below the suggested 25M. 
We evaluate our method across four ProcGen environments, each presenting unique and challenging properties, namely Caveflyer (open-world navigation with sparse rewards), Chaser (grid-based game with highly dense rewards), CoinRun (left-to-right platformer with highly sparse rewards), and Plunder (war game with dense rewards).

After training the world model and the agent on the collected steps, we keep on training the agent using dream-like trajectories from the fixed world model for another 1M steps. All the implementation details of the proposed solution in terms of architecture and training setup are reported in the \nameref{suppmaterial}.

\subsection{Baselines}

\begin{figure}[ht]
  \centering
  \includegraphics[width=1.\linewidth]{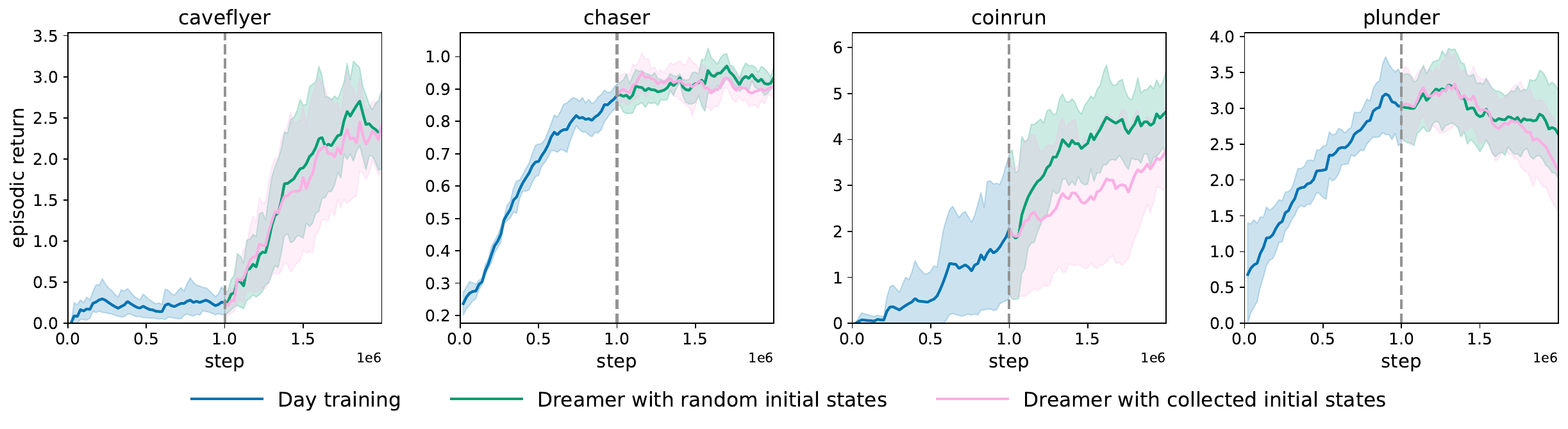}
  \caption[Impact of random initial states]{Total rewards received on all possible levels by classic Dreamer varying the source of initial states for imagination (randomly generated or collected from real environments). The vertical line separates the \textit{day} training (common to all methods) from the \textit{night} training. Results report average and confidence intervals across 5 seeds.}
  \label{fig:results_initial_states}
\end{figure}

We compare three variants of our method (which we refer to as RndDreamer, DeepDreamer, and ValDreamer since they use \textit{random swing}, \textit{DeepDream}, and \textit{value diversification} respectively) with two baselines: Dreamer-like training and Offline training. In this way, we study whether dream-like trajectories can improve generalization performances against classic imagination (without any transformation) and against further offline training over collected real experience. For a fair comparison, we use the same hyperparameters and use an offline adaptation of PPO \citep{queeney21} for offline training. Note that, unlike the original implementation, our Dreamer baseline considers randomly generated initial states instead of collected ones. As reported in Figure \ref{fig:results_initial_states}, we find that it helps obtain better generalization scores even without any further transformation.

\subsection{Results}

\begin{figure}[ht]
  \centering
  \includegraphics[width=1.\linewidth]{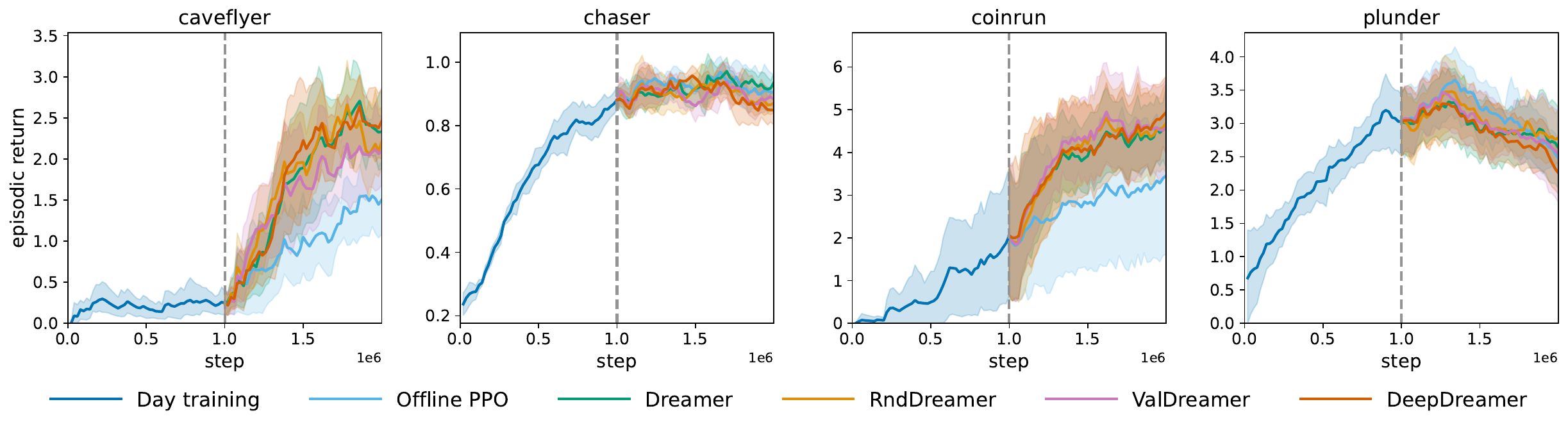}
  \caption[Generalization performances]{Total rewards received on all possible levels by our variants and by the two baselines. The vertical line separates the \textit{day} training (common to all methods) from the \textit{night} training. Results report average and confidence intervals across 5 seeds.}
  \label{fig:results}
\end{figure}

Figure \ref{fig:results} summarizes the results for all the four environments. As far as sparsely rewarded environments are concerned (i.e., Caveflyer and Coinrun), our variants consistently increase the rewards received by the agent, showing how the \textit{night} training is crucial to complement the \textit{day} training (reported in \textcolor{bluechart}{blue}). Offline training (in \textcolor{lightbluechart}{light blue}) only provides around 50\% of the improvement; while the variants of our proposed solution exceed standard imagination with random initial states (in \textcolor{greenchart}{green}) by a very small margin. This again proves the importance of starting from generated initial states (see Figure \ref{fig:results_initial_states}). On the contrary, results from densely rewarded environments (i.e., Chaser and Plunder) suggest that our method and in general imagination are of little help and can even cause catastrophic forgetting. Such different performances suggest that dream-like imagination is well-suited to complement the scarce information provided by a sparse environment when a limited amount of real experience is available. We also experiment with a mixture of the generative transformations without observing any further improvements (full results are reported in the \nameref{suppmaterial}).

\section{Conclusion} \label{conclusions}


In this paper, we have introduced a method for improving generalization in RL agents in case of limited training experience. Inspired by the \textit{Overfitted Brain} hypothesis, we have proposed to augment agent training through \textit{dream-like} imagination. In particular, we have discussed a method based on generating diverse imagined trajectories starting from randomly generated latent states and modifying intermediate ones with a set of state transformations. Our method has demonstrated superior generalization capabilities when compared to traditional imagination-based and offline RL techniques, particularly in scenarios characterized by very sparse rewards.  Our research agenda encompasses analyzing the scalability of our approach to tackle increasingly complex and diverse environments, as well as devising and assessing additional methods to generate even richer and more informative dream-like experiences.



\bibliography{main}

\begin{thebibliography}{63}
\providecommand{\natexlab}[1]{#1}
\providecommand{\url}[1]{\texttt{#1}}
\expandafter\ifx\csname urlstyle\endcsname\relax
  \providecommand{\doi}[1]{doi: #1}\else
  \providecommand{\doi}{doi: \begingroup \urlstyle{rm}\Url}\fi

\bibitem[Agarwal et~al.(2021)Agarwal, Machado, Castro, and
  Bellemare]{agarwal21}
Rishabh Agarwal, Marlos~C. Machado, Pablo~Samuel Castro, and Marc~G. Bellemare.
\newblock Contrastive behavioral similarity embeddings for generalization in
  reinforcement learning, 2021.
\newblock arXiv:2101.05265 [cs.LG].

\bibitem[Andrillon et~al.(2015)Andrillon, Nir, Cirelli, Tononi, and
  Fried]{andrillon15}
Thomas Andrillon, Yuval Nir, Chiara Cirelli, Giulio Tononi, and Itzhak Fried.
\newblock {Single-neuron activity and eye movements during human REM sleep and
  awake vision}.
\newblock \emph{Nature Communications}, 6\penalty0 (1):\penalty0 7884, 2015.

\bibitem[Banijamali et~al.(2018)Banijamali, Shu, Ghavamzadeh, Bui, and
  Ghodsi]{banijamali18}
Ershad Banijamali, Rui Shu, mohammad Ghavamzadeh, Hung Bui, and Ali Ghodsi.
\newblock Robust locally-linear controllable embedding.
\newblock In \emph{{Proceedings of the 21st International Conference on
  Artificial Intelligence and Statistics (AISTATS'18)}}, 2018.

\bibitem[Buesing et~al.(2018)Buesing, Weber, Racaniere, Eslami, Rezende,
  Reichert, Viola, Besse, Gregor, Hassabis, and Wierstra]{buesing18}
Lars Buesing, Theophane Weber, Sebastien Racaniere, S.~M.~Ali Eslami, Danilo
  Rezende, David~P. Reichert, Fabio Viola, Frederic Besse, Karol Gregor, Demis
  Hassabis, and Daan Wierstra.
\newblock Learning and querying fast generative models for reinforcement
  learning, 2018.
\newblock arXiv:1802.03006 [cs.LG].

\bibitem[Chen et~al.(2022)Chen, Wu, Yoon, and Ahn]{chen22}
Chang Chen, Yi-Fu Wu, Jaesik Yoon, and Sungjin Ahn.
\newblock {TransDreamer}: Reinforcement learning with transformer world models,
  2022.
\newblock arXiv:2202.09481 [cs.LG].

\bibitem[Cho et~al.(2014)Cho, van Merrienboer, Bahdanau, and Bengio]{cho14}
Kyunghyun Cho, Bart van Merrienboer, Dzmitry Bahdanau, and Yoshua Bengio.
\newblock On the properties of neural machine translation: Encoder-decoder
  approaches.
\newblock In \emph{{Proceedings of the 8th Workshop on Syntax, Semantics and
  Structure in Statistical Translation (SSST-8)}}, 2014.

\bibitem[Chua et~al.(2018)Chua, Calandra, McAllister, and Levine]{chua18}
Kurtland Chua, Roberto Calandra, Rowan McAllister, and Sergey Levine.
\newblock Deep reinforcement learning in a handful of trials using
  probabilistic dynamics models.
\newblock In \emph{{Advances in Neural Information Processing Systems
  (NIPS'18)}}, 2018.

\bibitem[Cobbe et~al.(2019)Cobbe, Klimov, Hesse, Kim, and Schulman]{cobbe19}
Karl Cobbe, Oleg Klimov, Chris Hesse, Taehoon Kim, and John Schulman.
\newblock Quantifying generalization in reinforcement learning.
\newblock In \emph{{Proceedings of the 36th International Conference on Machine
  Learning (ICML'19)}}, 2019.

\bibitem[Cobbe et~al.(2020)Cobbe, Hesse, Hilton, and Schulman]{cobbe20}
Karl Cobbe, Christopher Hesse, Jacob Hilton, and John Schulman.
\newblock Leveraging procedural generation to benchmark reinforcement learning.
\newblock In \emph{{Proceedings of the 37th International Conference on Machine
  Learning (ICML'20)}}, 2020.

\bibitem[Farebrother et~al.(2018)Farebrother, Machado, and
  Bowling]{farebrother18}
Jesse Farebrother, Marlos~C. Machado, and Michael Bowling.
\newblock Generalization and regularization in dqn, 2018.
\newblock arXiv:1810.00123 [cs.LG].

\bibitem[Gal et~al.(2016)Gal, McAllister, and Rasmussen]{gal16}
Yarin Gal, Rowan McAllister, and Carl~Edward Rasmussen.
\newblock Improving {PILCO} with {B}ayesian neural network dynamics models.
\newblock In \emph{{ICML'16 Workshop in Data-Efficient Machine Learning}},
  2016.

\bibitem[Ghosh et~al.(2021)Ghosh, Rahme, Kumar, Zhang, Adams, and
  Levine]{ghosh21}
Dibya Ghosh, Jad Rahme, Aviral Kumar, Amy Zhang, Ryan~P Adams, and Sergey
  Levine.
\newblock Why generalization in {RL} is difficult: Epistemic {POMDP}s and
  implicit partial observability.
\newblock In \emph{{Advances in Neural Information Processing Systems
  (NeurIPS'21)}}, 2021.

\bibitem[Ha \& Schmidhuber(2018)Ha and Schmidhuber]{ha18}
David Ha and J\"{u}rgen Schmidhuber.
\newblock Recurrent world models facilitate policy evolution.
\newblock In \emph{{Advances in Neural Information Processing Systems
  (NIPS'18)}}, 2018.

\bibitem[Hafner et~al.(2019)Hafner, Lillicrap, Fischer, Villegas, Ha, Lee, and
  Davidson]{hafner19}
Danijar Hafner, Timothy Lillicrap, Ian Fischer, Ruben Villegas, David Ha,
  Honglak Lee, and James Davidson.
\newblock Learning latent dynamics for planning from pixels.
\newblock In \emph{{Proceedings of the 36th International Conference on Machine
  Learning (ICML'19)}}, 2019.

\bibitem[Hafner et~al.(2020)Hafner, Lillicrap, Ba, and Norouzi]{hafner20a}
Danijar Hafner, Timothy Lillicrap, Jimmy Ba, and Mohammad Norouzi.
\newblock {Dream to Control: Learning Behaviors by Latent Imagination}.
\newblock In \emph{{Proceedings of the 8th International Conference on Learning
  Representations (ICLR'20)}}, 2020.

\bibitem[Hafner et~al.(2021)Hafner, Lillicrap, Norouzi, and Ba]{hafner21}
Danijar Hafner, Timothy Lillicrap, Mohammad Norouzi, and Jimmy Ba.
\newblock {Mastering Atari with Discrete World Models}.
\newblock In \emph{{Proceedings of the 9th International Conference on Learning
  Representations (ICLR'21)}}, 2021.

\bibitem[Hafner et~al.(2023)Hafner, Pasukonis, Ba, and Lillicrap]{hafner23}
Danijar Hafner, Jurgis Pasukonis, Jimmy Ba, and Timothy Lillicrap.
\newblock {Mastering Diverse Domains through World Models}, 2023.
\newblock arXiv:2301.04104 [cs.AI].

\bibitem[Henaff et~al.(2017)Henaff, Whitney, and LeCun]{henaff17}
Mikael Henaff, William~F. Whitney, and Yann LeCun.
\newblock Model-based planning with discrete and continuous actions, 2017.
\newblock arXiv:1705.07177 [cs.AI].

\bibitem[Hoel(2019)]{hoel19}
Erik Hoel.
\newblock Enter the supersensorium: The neuroscientific case for art in the age
  of netflix.
\newblock \emph{The Baffler}, 45, 2019.
\newblock \url{https://thebaffler.com/salvos/enter-the-supersensorium-hoel}.

\bibitem[Hoel(2021)]{hoel21}
Erik Hoel.
\newblock The overfitted brain: Dreams evolved to assist generalization.
\newblock \emph{Patterns}, 2\penalty0 (5):\penalty0 100244, 2021.

\bibitem[Igl et~al.(2018)Igl, Zintgraf, Le, Wood, and Whiteson]{igl18}
Maximilian Igl, Luisa Zintgraf, Tuan~Anh Le, Frank Wood, and Shimon Whiteson.
\newblock Deep variational reinforcement learning for {POMDP}s.
\newblock In \emph{{Proceedings of the 35th International Conference on Machine
  Learning (ICML'18)}}, 2018.

\bibitem[Igl et~al.(2019)Igl, Ciosek, Li, Tschiatschek, Zhang, Devlin, and
  Hofmann]{igl19}
Maximilian Igl, Kamil Ciosek, Yingzhen Li, Sebastian Tschiatschek, Cheng Zhang,
  Sam Devlin, and Katja Hofmann.
\newblock Generalization in reinforcement learning with selective noise
  injection and information bottleneck.
\newblock In \emph{{Advances in Neural Information Processing Systems
  (NeurIPS'19)}}, 2019.

\bibitem[Jaderberg et~al.(2017)Jaderberg, Mnih, Czarnecki, Schaul, Leibo,
  Silver, and Kavukcuoglu]{jaderberg17}
Max Jaderberg, Volodymyr Mnih, Wojciech~Marian Czarnecki, Tom Schaul, Joel~Z.
  Leibo, David Silver, and Koray Kavukcuoglu.
\newblock Reinforcement learning with unsupervised auxiliary tasks.
\newblock In \emph{{Proceedings of the 5th International Conference on Learning
  Representations (ICLR'17)}}, 2017.

\bibitem[Jiang et~al.(2021)Jiang, Grefenstette, and Rockt{\"a}schel]{jiang21}
Minqi Jiang, Edward Grefenstette, and Tim Rockt{\"a}schel.
\newblock Prioritized level replay.
\newblock In \emph{{Proceedings of the 38th International Conference on Machine
  Learning (ICML'21)}}, 2021.

\bibitem[Kaiser et~al.(2020)Kaiser, Babaeizadeh, Milos, Osinski, Campbell,
  Czechowski, Erhan, Finn, Kozakowski, Levine, Mohiuddin, Sepassi, Tucker, and
  Michalewski]{kaiser20}
Lukasz Kaiser, Mohammad Babaeizadeh, Piotr Milos, Blazej Osinski, Roy~H.
  Campbell, Konrad Czechowski, Dumitru Erhan, Chelsea Finn, Piotr Kozakowski,
  Sergey Levine, Afroz Mohiuddin, Ryan Sepassi, George Tucker, and Henryk
  Michalewski.
\newblock Model based reinforcement learning for atari.
\newblock In \emph{{Proceedings of the 8th International Conference on Learning
  Representations (ICLR'20)}}, 2020.

\bibitem[Kingma et~al.(2016)Kingma, Salimans, Jozefowicz, Chen, Sutskever, and
  Welling]{kingma16}
Durk~P Kingma, Tim Salimans, Rafal Jozefowicz, Xi~Chen, Ilya Sutskever, and Max
  Welling.
\newblock Improved variational inference with inverse autoregressive flow.
\newblock In \emph{{Advances in Neural Information Processing Systems
  (NIPS'16)}}, 2016.

\bibitem[Kirk et~al.(2023)Kirk, Zhang, Grefenstette, and
  Rockt\"{a}schel]{kirk23}
Robert Kirk, Amy Zhang, Edward Grefenstette, and Tim Rockt\"{a}schel.
\newblock A survey of zero-shot generalisation in deep reinforcement learning.
\newblock \emph{Journal of Artificial Intelligence Research}, 76:\penalty0 64,
  2023.

\bibitem[Laskin et~al.(2020)Laskin, Lee, Stooke, Pinto, Abbeel, and
  Srinivas]{laskin20}
Misha Laskin, Kimin Lee, Adam Stooke, Lerrel Pinto, Pieter Abbeel, and Aravind
  Srinivas.
\newblock Reinforcement learning with augmented data.
\newblock In \emph{{Advances in Neural Information Processing Systems
  (NeurIPS'20)}}, 2020.

\bibitem[Lee et~al.(2020{\natexlab{a}})Lee, Nagabandi, Abbeel, and
  Levine]{lee20a}
Alex~X. Lee, Anusha Nagabandi, Pieter Abbeel, and Sergey Levine.
\newblock Stochastic latent actor-critic: Deep reinforcement learning with a
  latent variable model.
\newblock In \emph{{Advances in Neural Information Processing Systems
  (NeurIPS'20)}}, 2020{\natexlab{a}}.

\bibitem[Lee et~al.(2020{\natexlab{b}})Lee, Lee, Shin, and Lee]{lee20b}
Kimin Lee, Kibok Lee, Jinwoo Shin, and Honglak Lee.
\newblock Network randomization: {A} simple technique for generalization in
  deep reinforcement learning.
\newblock In \emph{{Proceedings of the 8th International Conference on Learning
  Representations (ICLR'20)}}, 2020{\natexlab{b}}.

\bibitem[Lillicrap et~al.(2016)Lillicrap, Hunt, Pritzel, Heess, Erez, Tassa,
  Silver, and Wierstra]{lillicrap16}
Timothy~P. Lillicrap, Jonathan~J. Hunt, Alexander Pritzel, Nicolas Heess, Tom
  Erez, Yuval Tassa, David Silver, and Daan Wierstra.
\newblock Continuous control with deep reinforcement learning.
\newblock In \emph{{Proceedings of the 4th International Conference on Learning
  Representations (ICLR'16)}}, 2016.

\bibitem[Lyle et~al.(2022)Lyle, Rowland, Dabney, Kwiatkowska, and Gal]{lyle22}
Clare Lyle, Mark Rowland, Will Dabney, Marta Kwiatkowska, and Yarin Gal.
\newblock Learning dynamics and generalization in deep reinforcement learning.
\newblock In \emph{{Proceedings of the 39th International Conference on Machine
  Learning (ICML'22)}}, 2022.

\bibitem[Micheli et~al.(2023)Micheli, Alonso, and Fleuret]{micheli23}
Vincent Micheli, Eloi Alonso, and Fran{\c{c}}ois Fleuret.
\newblock Transformers are sample-efficient world models.
\newblock In \emph{{Proceedings of the 11th International Conference on
  Learning Representations (ICLR'23)}}, 2023.

\bibitem[Micikevicius et~al.(2018)Micikevicius, Narang, Alben, Diamos, Elsen,
  Garcia, Ginsburg, Houston, Kuchaiev, Venkatesh, and Wu]{micikevicius18}
Paulius Micikevicius, Sharan Narang, Jonah Alben, Gregory Diamos, Erich Elsen,
  David Garcia, Boris Ginsburg, Michael Houston, Oleksii Kuchaiev, Ganesh
  Venkatesh, and Hao Wu.
\newblock Mixed precision training.
\newblock In \emph{{Proceedings of the 6th International Conference on Learning
  Representations (ICLR'18)}}, 2018.

\bibitem[Mnih et~al.(2015)Mnih, Kavukcuoglu, Silver, Rusu, Veness, Bellemare,
  Graves, Riedmiller, Fidjeland, Ostrovski, Petersen, Beattie, Sadik,
  Antonoglou, King, Kumaran, Wierstra, Legg, and Hassabis]{mnih15}
Volodymyr Mnih, Koray Kavukcuoglu, David Silver, Andrei~A. Rusu, Joel Veness,
  Marc~G. Bellemare, Alex Graves, Martin Riedmiller, Andreas~K. Fidjeland,
  Georg Ostrovski, Stig Petersen, Charles Beattie, Amir Sadik, Ioannis
  Antonoglou, Helen King, Dharshan Kumaran, Daan Wierstra, Shane Legg, and
  Demis Hassabis.
\newblock Human-level control through deep reinforcement learning.
\newblock \emph{Nature}, 518\penalty0 (7540):\penalty0 529--533, 2015.

\bibitem[Mordvintsev et~al.(2015)Mordvintsev, Olah, and Tyka]{mordvintsev15}
Alexander Mordvintsev, Christopher Olah, and Mike Tyka.
\newblock Inceptionism: Going deeper into neural networks, 2015.
\newblock Google Research Blog.

\bibitem[Mu et~al.(2021)Mu, Zhuang, Wang, Zhu, Liu, Chen, Luo, Li, Zhang, and
  Hao]{mu21}
Yao Mu, Yuzheng Zhuang, Bin Wang, Guangxiang Zhu, Wulong Liu, Jianyu Chen, Ping
  Luo, Shengbo Li, Chongjie Zhang, and Jianye Hao.
\newblock Model-based reinforcement learning via imagination with derived
  memory.
\newblock In \emph{{Advances in Neural Information Processing Systems
  (NeurIPS'21)}}, 2021.

\bibitem[Nagabandi et~al.(2018)Nagabandi, Kahn, Fearing, and
  Levine]{nagabandi18}
Anusha Nagabandi, Gregory Kahn, Ronald~S. Fearing, and Sergey Levine.
\newblock Neural network dynamics for model-based deep reinforcement learning
  with model-free fine-tuning.
\newblock In \emph{{Proceedings of the 2018 IEEE International Conference on
  Robotics and Automation (ICRA'18)}}, 2018.

\bibitem[Ouyang et~al.(2022)Ouyang, Wu, Jiang, Almeida, Wainwright, Mishkin,
  Zhang, Agarwal, Slama, Ray, Schulman, Hilton, Kelton, Miller, Simens, Askell,
  Welinder, Christiano, Leike, and Lowe]{ouyang22}
Long Ouyang, Jeffrey Wu, Xu~Jiang, Diogo Almeida, Carroll Wainwright, Pamela
  Mishkin, Chong Zhang, Sandhini Agarwal, Katarina Slama, Alex Ray, John
  Schulman, Jacob Hilton, Fraser Kelton, Luke Miller, Maddie Simens, Amanda
  Askell, Peter Welinder, Paul~F Christiano, Jan Leike, and Ryan Lowe.
\newblock Training language models to follow instructions with human feedback.
\newblock In \emph{{Advances in Neural Information Processing Systems
  (NeurIPS'22)}}, 2022.

\bibitem[Queeney et~al.(2021)Queeney, Paschalidis, and Cassandras]{queeney21}
James Queeney, Ioannis Paschalidis, and Christos Cassandras.
\newblock Generalized proximal policy optimization with sample reuse.
\newblock In \emph{{Advances in Neural Information Processing Systems
  (NeurIPS'21)}}, 2021.

\bibitem[Racani\`{e}re et~al.(2017)Racani\`{e}re, Weber, Reichert, Buesing,
  Guez, Jimenez~Rezende, Puigdom\`{e}nech~Badia, Vinyals, Heess, Li, Pascanu,
  Battaglia, Hassabis, Silver, and Wierstra]{racaniere17}
S\'{e}bastien Racani\`{e}re, Theophane Weber, David Reichert, Lars Buesing,
  Arthur Guez, Danilo Jimenez~Rezende, Adri\`{a} Puigdom\`{e}nech~Badia, Oriol
  Vinyals, Nicolas Heess, Yujia Li, Razvan Pascanu, Peter Battaglia, Demis
  Hassabis, David Silver, and Daan Wierstra.
\newblock Imagination-augmented agents for deep reinforcement learning.
\newblock In \emph{{Advances in Neural Information Processing Systems
  (NIPS'17)}}, 2017.

\bibitem[Raileanu \& Fergus(2021)Raileanu and Fergus]{raileanu21b}
Roberta Raileanu and Rob Fergus.
\newblock Decoupling value and policy for generalization in reinforcement
  learning.
\newblock In \emph{{Proceedings of the 38th International Conference on Machine
  Learning (ICML'21)}}, 2021.

\bibitem[Raileanu et~al.(2021)Raileanu, Goldstein, Yarats, Kostrikov, and
  Fergus]{raileanu21a}
Roberta Raileanu, Maxwell Goldstein, Denis Yarats, Ilya Kostrikov, and Rob
  Fergus.
\newblock Automatic data augmentation for generalization in reinforcement
  learning.
\newblock In \emph{{Advances in Neural Information Processing Systems
  (NeurIPS'21)}}, 2021.

\bibitem[Revonsuo(2000)]{revonsuo00}
Antti Revonsuo.
\newblock The reinterpretation of dreams: An evolutionary hypothesis of the
  function of dreaming.
\newblock \emph{Behavioral and Brain Sciences}, 23\penalty0 (6):\penalty0
  877–901, 2000.

\bibitem[Schrittwieser et~al.(2020)Schrittwieser, Antonoglou, Hubert, Simonyan,
  Sifre, Schmitt, Guez, Lockhart, Hassabis, Graepel, Lillicrap, and
  Silver]{schrittwieser20}
Julian Schrittwieser, Ioannis Antonoglou, Thomas Hubert, Karen Simonyan,
  Laurent Sifre, Simon Schmitt, Arthur Guez, Edward Lockhart, Demis Hassabis,
  Thore Graepel, Timothy Lillicrap, and David Silver.
\newblock {Mastering Atari, Go, chess and shogi by planning with a learned
  model}.
\newblock \emph{Nature}, 588:\penalty0 604--609, 2020.

\bibitem[Schulman et~al.(2016)Schulman, Moritz, Levine, Jordan, and
  Abbeel]{schulman15}
John Schulman, Philipp Moritz, Sergey Levine, Michael~I. Jordan, and Pieter
  Abbeel.
\newblock High-dimensional continuous control using generalized advantage
  estimation.
\newblock In \emph{{Proceedings of the 4th International Conference on Learning
  Representations (ICLR'16)}}, 2016.

\bibitem[Schulman et~al.(2017)Schulman, Wolski, Dhariwal, Radford, and
  Klimov]{schulman17}
John Schulman, Filip Wolski, Prafulla Dhariwal, Alec Radford, and Oleg Klimov.
\newblock Proximal policy optimization algorithms, 2017.
\newblock arXiv:1707.06347 [cs.LG].

\bibitem[Sekar et~al.(2020)Sekar, Rybkin, Daniilidis, Abbeel, Hafner, and
  Pathak]{sekar20}
Ramanan Sekar, Oleh Rybkin, Kostas Daniilidis, Pieter Abbeel, Danijar Hafner,
  and Deepak Pathak.
\newblock Planning to explore via self-supervised world models.
\newblock In \emph{{Proceedings of the 37th International Conference on Machine
  Learning (ICML'20)}}, 2020.

\bibitem[Simonyan et~al.(2014)Simonyan, Vedaldi, and Zisserman]{simonyan14}
Karen Simonyan, Andrea Vedaldi, and Andrew Zisserman.
\newblock Deep inside convolutional networks: Visualising image classification
  models and saliency maps.
\newblock In \emph{{Proceedings of the ICLR'14 Workshop}}, 2014.

\bibitem[Stooke et~al.(2021)Stooke, Lee, Abbeel, and Laskin]{stooke21}
Adam Stooke, Kimin Lee, Pieter Abbeel, and Michael Laskin.
\newblock Decoupling representation learning from reinforcement learning.
\newblock In \emph{{Proceedings of the 38th International Conference on Machine
  Learning (ICML'21)}}, 2021.

\bibitem[Sutton \& Barto(2017)Sutton and Barto]{sutton17}
R.~S. Sutton and A.~G. Barto.
\newblock \emph{{Reinforcement Learning: An Introduction}}.
\newblock The MIT Press, 2017.

\bibitem[Sutton(1991)]{sutton91}
Richard~S. Sutton.
\newblock Dyna, an integrated architecture for learning, planning, and
  reacting.
\newblock In \emph{{Working Notes of the 1991 AAAI Spring Symposium}}, 1991.

\bibitem[Tobin et~al.(2017)Tobin, Fong, Ray, Schneider, Zaremba, and
  Abbeel]{tobin17}
Josh Tobin, Rachel Fong, Alex Ray, Jonas Schneider, Wojciech Zaremba, and
  Pieter Abbeel.
\newblock Domain randomization for transferring deep neural networks from
  simulation to the real world.
\newblock In \emph{{Proceedings of the 2017 IEEE/RSJ International Conference
  on Intelligent Robots and Systems (IROS'17)}}, 2017.

\bibitem[Tsividis et~al.(2017)Tsividis, Pouncy, Xu, Tenenbaum, and
  Gershman]{tsividis17}
Pedro Tsividis, Thomas Pouncy, Jaqueline~L. Xu, Joshua~B. Tenenbaum, and
  Samuel~J. Gershman.
\newblock Human learning in atari.
\newblock In \emph{{Proceedings of the 2017 AAAI Spring Symposium Series,
  Science of Intelligence: Computational Principles of Natural and Artificial
  Intelligence}}, 2017.

\bibitem[Vaswani et~al.(2017)Vaswani, Shazeer, Parmar, Uszkoreit, Jones, Gomez,
  Kaiser, and Polosukhin]{vaswani17}
Ashish Vaswani, Noam Shazeer, Niki Parmar, Jakob Uszkoreit, Llion Jones,
  Aidan~N. Gomez, Lukasz Kaiser, and Illia Polosukhin.
\newblock Attention is all you need.
\newblock In \emph{{Advances in Neural Information Processing Systems
  (NIPS'17)}}, 2017.

\bibitem[Wang et~al.(2020)Wang, Kang, Shao, and Feng]{wang20}
Kaixin Wang, Bingyi Kang, Jie Shao, and Jiashi Feng.
\newblock Improving generalization in reinforcement learning with mixture
  regularization.
\newblock In \emph{{Advances in Neural Information Processing Systems
  (NeurIPS'20)}}, 2020.

\bibitem[Watter et~al.(2015)Watter, Springenberg, Boedecker, and
  Riedmiller]{watter15}
Manuel Watter, Jost Springenberg, Joschka Boedecker, and Martin Riedmiller.
\newblock Embed to control: A locally linear latent dynamics model for control
  from raw images.
\newblock In \emph{{Advances in Neural Information Processing Systems
  (NIPS'15)}}, 2015.

\bibitem[Webber(2012)]{webber13}
J~Beau~W Webber.
\newblock A bi-symmetric log transformation for wide-range data.
\newblock \emph{Measurement Science and Technology}, 24\penalty0 (2):\penalty0
  027001, 2012.

\bibitem[Williams(1992)]{williams92}
Ronald~J. Williams.
\newblock Simple statistical gradient-following algorithms for connectionist
  reinforcement learning.
\newblock \emph{Machine Learning}, 8:\penalty0 229--256, 1992.

\bibitem[Yarats et~al.(2021)Yarats, Kostrikov, and Fergus]{yarats21}
Denis Yarats, Ilya Kostrikov, and Rob Fergus.
\newblock Image augmentation is all you need: Regularizing deep reinforcement
  learning from pixels.
\newblock In \emph{{Proceedings of the 9th International Conference on Learning
  Representations (ICLR'21)}}, 2021.

\bibitem[Ye et~al.(2020)Ye, Khalifa, Bontrager, and Togelius]{ye20}
Chang Ye, Ahmed Khalifa, Philip Bontrager, and Julian Togelius.
\newblock Rotation, translation, and cropping for zero-shot generalization,
  2020.
\newblock arXiv:2001.09908 [cs.LG].

\bibitem[Zhang et~al.(2019)Zhang, Vikram, Smith, Abbeel, Johnson, and
  Levine]{zhang19}
Marvin Zhang, Sharad Vikram, Laura~M. Smith, Pieter Abbeel, Matthew~J. Johnson,
  and Sergey Levine.
\newblock {SOLAR:} deep structured representations for model-based
  reinforcement learning.
\newblock In \emph{{Proceedings of the 36th International Conference on Machine
  Learning (ICML'19)}}, 2019.

\bibitem[Zhu et~al.(2020)Zhu, Zhang, Lee, and Zhang]{zhu20}
Guangxiang Zhu, Minghao Zhang, Honglak Lee, and Chongjie Zhang.
\newblock Bridging imagination and reality for model-based deep reinforcement
  learning.
\newblock In \emph{{Advances in Neural Information Processing Systems
  (NeurIPS'20)}}, 2020.

\end{thebibliography}
\bibliographystyle{rlc}

\newpage
\appendix

\section*{Supplementary Material} \label{suppmaterial}

\subsection*{Implementation Details for Reproducibility}

\label{sec:parameters}
We adopt small model sizes for our neural networks as in \cite{hafner23}; Table \ref{tab:modelparams} reports the full list of network hyperparameters. Table \ref{tab:expparams} reports all the training parameters. In addition, we also leverage mixed precision \citep{micikevicius18} to reduce resource consumption. 

\begin{table}[ht]
    \centering    
    \begin{tabular}{lc}
        \hline \noalign{\vskip 1mm}
        \textbf{Parameter} & \textbf{Value} \\ [0.5ex]
        \hline
        Categoricals $C$ & 32 \\
        Classes $J$ & 32 \\
        RNN hidden units & 512 \\
        Convolution filters & [32, 64, 128, 256] \\
        Convolution kernel size & 4 \\
        Convolution strides & 2 \\
        Deconvolution filters & [128, 64, 32, 3] \\
        Deconvolution kernel size & 4 \\
        Deconvolution strides & 2 \\
        Linear units & 512 \\
        MLP layers & 2 \\
        Normalization & Layer \\
        Activation & swish \\
        Learning rate during \textit{day} & 5e-4 \\
        Learning rate at \textit{night} & 1e-4 \\
        Optimizer & Adam \\
        Reward/return bins $K$ & 255 \\
        Bins extremes & -20, +20 \\
        Dynamics loss factor $\beta_1$ & 0.5 \\
        Representation loss factor $\beta_2$ & 0.1 \\
        Critic loss factor $c_v$ & 0.5 \\
        Entropy loss factor $c_e$ & 0.001 \\
        $\gamma$ parameter during \textit{day} (GAE) & 0.99 \\
        $\gamma$ parameter at \textit{night} (GAE) & 1 - 1/H \\
        $\lambda$ parameter (GAE) & 0.95 \\
        PPO clip factor $\epsilon$ & 0.2 \\
        PPO gradient clip factor & 0.5 \\
        PPO iterations & 4 \\      
    \end{tabular}
    \caption{Network hyperparameters.}
    \label{tab:modelparams}
\end{table}

\begin{table}[ht]
    \centering    
    \begin{tabular}{lc}
        \hline \noalign{\vskip 1mm}
        \textbf{Parameter} & \textbf{Value} \\ [0.5ex]
        \hline
        Seed episodes $S$ & 5 \\
        \textit{Day} epochs $E_d$ & 200 \\
        World model update steps $U_w$ & 20 \\
        \textit{Day} steps per epoch $T_d$ & 5000 \\
        World batch size $B_w$ & 100 \\
        Sequence length $L$ & 25 \\
        \textit{Night} epochs $E_n$ & 200 \\
        Imagination update steps $U_a$ & 26 \\
        Imagination batch size $B_a$ & 12 \\
        Imagination horizon $H$ & 16 \\
        Test repetition & 5 \\
        Parallelized environments & 5 \\
        Penalty for non-successful episodes (coinrun and caveflyer) & -10 \\
        Penalty for non-successful episodes (chaser and plunder) & 0 \\
        Reward scaling factor (chaser only) & 25 \\
        DeepDream optimization steps & 10 \\
        DeepDream step size & 0.1 \\
        Value maximization optimization steps & 10 \\
        Value maximization step size & 0.5 \\
    \end{tabular}
    \caption{Training parameters.}
    \label{tab:expparams}
\end{table}

\clearpage
\appendix

\subsection*{Additional Results}

\begin{figure}[ht]
  \centering
  \includegraphics[width=1.\linewidth]{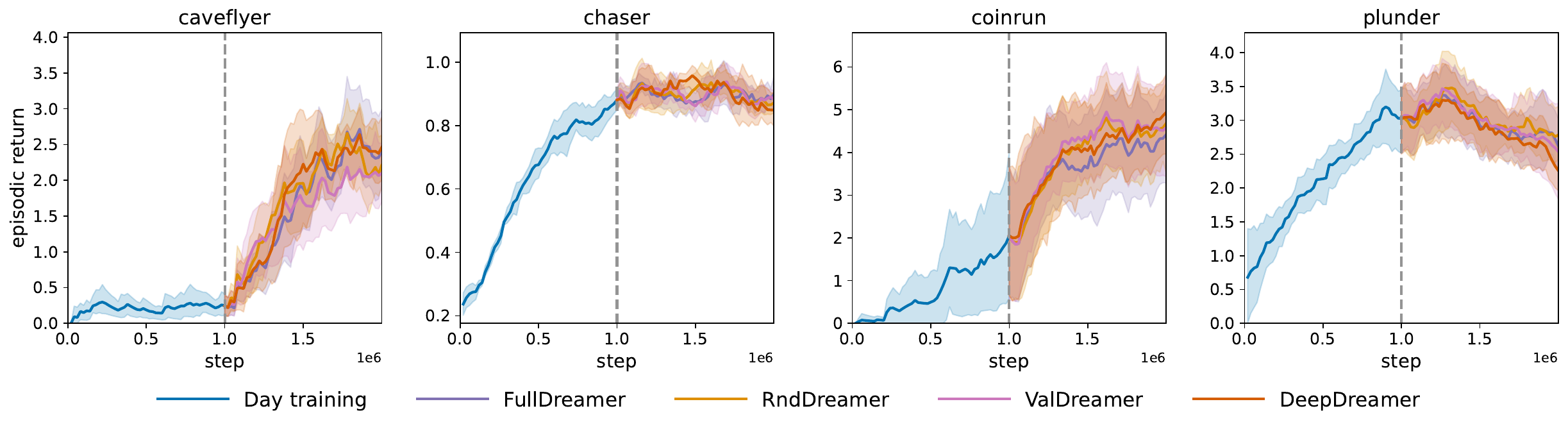}
  \caption[Transformations performances]{Total rewards received on all the levels by our variants considering the transformations separately and together with random uniform probability. The vertical line separates the \textit{day} training (common to all methods) from the \textit{night} training. Results report average and confidence intervals across 5 seeds.}
  \label{fig:all_results}
\end{figure}

In addition to the results presented in Section \ref{experiments}, we also experiment with a mixture of the generative transformations, i.e., by randomly applying the three transformations with equal probability to the current state. We refer to this variant as FullDreamer. As reported in Figure \ref{fig:all_results}, we do not observe any significant improvement compared with the cases in which the transformations are applied separately. This is probably due to the fact that one of them has a prominent effect on the learning performance.

\end{document}